\pdfoutput=1
\documentclass[11pt,a4paper]{article}
\usepackage[hyperref]{emnlp2020}
\usepackage{times}
\usepackage{latexsym}

\usepackage{microtype}

\usepackage{xspace}
\usepackage{url}
\usepackage{color}
\usepackage{xcolor}
\usepackage{multirow}
\usepackage{amsmath}
\usepackage{graphicx}
\usepackage{amsfonts}
\usepackage{colortbl}
\usepackage{booktabs}
\usepackage{textcomp}

\aclfinalcopy 

\newcommand{\nligen}{GenNLI\xspace}
\newcommand{\gennli}{\nligen}

\newcommand{\theloss}{infinilog\xspace}

\newcommand{\softmax}{\mathrm{softmax}}
\newcommand{\vocab}{\mathit{vocab}}

\newcommand{\mcopy}{\mathit{copy}}
\newcommand{\bos}{\mathit{BOS}}

\newcommand{\cost}{\mathrm{cost}}

\newcommand{\argmax}{\operatornamewithlimits{argmax}}

\newenvironment{itemizesquish}{\begin{list}{\labelitemi}{\setlength{\itemsep}{-0.2em}\setlength{\labelwidth}{0.5em}\setlength{\leftmargin}{\labelwidth}\addtolength{\leftmargin}{\labelsep}}}{\end{list}}

\title{Discriminatively-Tuned Generative Classifiers \\ for Robust Natural Language Inference}

\author{Xiaoan Ding$^1$ \Thanks{ Equal contribution.} \ \ \   Tianyu Liu$^3$ \footnotemark[1] \ \Thanks{ Contribution during visiting TTIC.} \ \ \ Baobao Chang$^3$ $^4$ \ \ \ Zhifang Sui$^3$ $^4$ \ \ \  Kevin Gimpel$^2$  \\
 $^1$ University of Chicago, IL, USA \ \ $^2$ Toyota Technological Institute at Chicago, IL, USA \\
 $^3$ Peking University, Beijing, China \ \  $^4$ Peng Cheng Laboratory, Shenzhen, China\\
 \texttt{xiaoanding@uchicago.edu,} \
 \texttt{\{tianyu0421, chbb, szf\}@pku.edu.cn,} \\
  \texttt{kgimpel@ttic.edu}  }

\date{}

\begin{document}
\maketitle
\begin{abstract}

While discriminative neural network classifiers are generally preferred, recent work has shown advantages of generative classifiers in term of data efficiency and robustness. In this paper, we focus on natural language inference (NLI). We propose \gennli, a generative classifier for NLI tasks, and empirically characterize its performance by comparing it to five baselines, including discriminative models and large-scale pretrained language representation models like BERT. 
We explore training objectives for discriminative fine-tuning of our generative classifiers, showing improvements over log loss fine-tuning from prior work~\citep{lewis2018generative}. In particular, we find strong results with a simple unbounded modification to log loss, which we call the ``\theloss loss''. 
Our experiments show that \gennli outperforms both discriminative and pretrained baselines across several challenging NLI experimental settings, including small training sets, imbalanced label distributions, and label noise.

\end{abstract}

\section{Introduction}
\pdfoutput=1
\label{intro}

Natural language inference (NLI) is the task of identifying the relationship between two fragments of text, called the  \emph{premise} and the \emph{hypothesis}~\citep{dagan2006pascal,DBLP:series/synthesis/2013Dagan}. The task was originally defined as binary classification, in which the labels are \emph{entailment} (the premise implies the hypothesis) or \emph{not entailment}. Subsequent variations added a third \emph{contradiction} label. Most models for NLI are trained and evaluated on standard benchmarks \citep{snli:emnlp2015,williams-etal-2018-broad,wang2018glue} in a discriminative manner \citep{conneau-etal-2017-supervised,chen-etal-2017-enhanced}. These benchmarks  typically have relatively clean, balanced, and abundant annotated data, and there is no distribution shift between the training and test sets.

However, when data quality and conditions are not ideal, there is a substantial performance decrease for  existing discriminative models, including both simple model architectures and more complex ones. 
Prior work on document classification and question answering has shown that \textbf{generative classifiers} have advantages over their discriminative counterparts in non-ideal conditions \citep{yogatama2017generative,lewis2018generative,ding-gimpel-2019-latent}. 

In this paper, we develop generative classifiers for NLI. Our model, which we call \gennli, defines the conditional probability of the hypothesis given the premise and the label, parameterizing the distribution using a sequence-to-sequence model with attention~\citep{luong-etal-2015-effective} and a copy mechanism~\citep{gu-etal-2016-incorporating}. We explore training objectives for discriminative fine-tuning of our generative classifiers, comparing several classical discriminative criteria. We find that several losses, including hinge loss and softmax-margin, outperform log loss fine-tuning used in prior work~\citep{lewis2018generative} while similarly retaining the advantages of generative classifiers. We also find strong results with a simple unbounded modification to log loss, which we call the ``\theloss loss''. 

Our evaluation focuses on challenging experimental conditions: small training sets, imbalanced label distributions, and label noise. 
We empirically compare \gennli with several discriminative baselines and large-scale pretrained language representation models~\citep{devlin-etal-2019-bert,yang2019xlnet,liu2019roberta} on five standard datasets. \gennli has better performance than discriminative classifiers under the small data setting. Moreover, when limited to 100 instances per class, \gennli consistently outperforms all BERT-style pretrained models on four of the five datasets. These results are appealing especially in comparison with BERT-style pretrained baselines.  
Large-scale pretrained language models have achieved state-of-the-art results on a wide range of NLP tasks, but they still require hundreds or even thousands of annotated examples to outperform \gennli. 

\gennli also outperforms discriminative classifiers when the training data shows severe label imbalance and when training labels are randomly corrupted. We additionally use \gennli to generate hypotheses for given premises and labels. 
While the generations tend to have low diversity due to high lexical overlap with the premise, they are generally fluent and comport with the given labels,  even in the small data setting.

\section{Background and Related Work}
\pdfoutput=1
\label{relatedwork}
\subsection{Generative Classifiers}
While discriminative classifiers directly model the posterior probability of the label given the input, i.e., $p(y\mid x)$, generative classifiers instead model the joint probability $p(x, y)$, typically factoring it into $p(x\mid y)$ and $p(y)$ and making decisions as follows:
\[ \hat{y} = \argmax_y p(x\mid y) p(y)\] 
Most neural network classifiers are trained as discriminative classifiers as these work better when conditions are favorable for supervised learning, namely that training data is plentiful and that the training and test data are drawn from the same distribution. While discriminative classifiers are generally preferred in practice, there is certain prior work showing that generative classifiers can have advantages in certain conditions, especially when training data is scarce, noisy, and imbalanced~\citep{yogatama2017generative,lewis2018generative,ding-gimpel-2019-latent}. 

\citet{ng2002discriminative} proved theoretically  that generative classifiers can approach their asymptotic error much faster, as na\"ive Bayes is faster than its discriminative analogue, logistic regression. \citet{yogatama2017generative} compared the performance of generative and discriminative classifiers and showed the advantages of neural generative classifiers in terms of sample complexity, data shift, and zero-shot and continual learning settings. \citet{ding-gimpel-2019-latent} further improved the performance of generative classifiers on document classification by introducing discrete latent variables into the generative story. \citet{lewis2018generative} developed generative classifiers for question answering and achieved comparable performance to discriminative models on the SQuAD~\citep{rajpurkar-etal-2016-squad} dataset, and much better performance in challenging experimental settings. 

In this paper, we develop generative models for natural language inference inspired by models for sequence-to-sequence tasks. We additionally contribute an exploration of several discriminative objectives for fine-tuning our generative classifiers, finding multiple choices to outperform log loss used in prior work. We also compare our generative classifiers with fine-tuning of large-scale pretrained models, and characterize performance under other realistic settings such as imbalanced and noisy datasets.

\subsection{Natural Language Inference}
 
Early methods for NLI mainly relied on conventional, feature-based methods trained from small-scale datasets \cite{DBLP:series/synthesis/2013Dagan,marelli-etal-2014-sick}. The release of larger datasets, such as SNLI, made neural network methods feasible. Such methods can be roughly categorized into two classes: sentence embedding bottleneck methods which first encode the two sentences as vectors and then feed them into a classifier for classification \cite{conneau-etal-2017-supervised,nie-bansal-2017-shortcut,choi2018learning,chen-etal-2017-recurrent-neural,wu2018phrase}, and more general methods which usually involve interactions while encoding the two sentences in the pair \cite{chen-etal-2017-enhanced,gong2018natural,parikh-etal-2016-decomposable}. Recently, NLI models are shown to be biased towards spurious surface patterns in the human annotated datasets \cite{poliak2018hypothesis,gururangan2018annotation,liu2020hyponli}, which makes them vulnerable to adversarial attacks \cite{GlocknerSG18,minervini2018adversarially,mccoy2019right,liu2020empirical}.

\section{A Generative Classifier for NLI}
\pdfoutput=1
\label{model}
Each example in a natural language inference dataset consists of two natural language texts, known as the premise and the hypothesis, and a label indicating the relation between the two texts. Formally, we denote an instance $\langle x^{(p)}, x^{(h)}, y \rangle $ as a tuple consisting of a premise $x^{(p)} = \{x^{(p)}_1, x^{(p)}_2, ..., x^{(p)}_N\}$, a hypothesis $x^{(h)} = \{x^{(h)}_1, x^{(h)}_2, ..., x^{(h)}_T\}$, and a label $y \in Y$. 

Most existing NLI models are trained in a discriminative manner by maximizing the conditional log-likelihood of the label given the input, i.e., $\log p(y \mid x^{(p)}, x^{(h)})$. In this paper, we propose generative classifiers for NLI that are trained instead to estimate the probability of the hypothesis given the premise and the label, i.e., $p(x^{(h)} \mid x^{(p)}, y)$, typically by maximizing log-likelihood. We decompose this conditional probability using the chain rule, and our final training objective is to minimize the following negative log likelihood:
\begin{align}
L(x^{(p)}, x^{(h)}, y) =  -\!\sum_{t=1}^{T} \log p(x^{(h)}_t \mid x^{(h)}_{<t}, x^{(p)}, y) \label{eq:train_objective}
\end{align}
At inference time, the prediction is made as follows:
\begin{align}
\argmax_{y \in Y} \,\log p(y) + \! \sum_{t=1}^{T}  \log  p(x^{(h)}_t \mid x^{(h)}_{<t}, x^{(p)}, y)  
\label{eq:predict}
\end{align}
Throughout all of the experiments in this paper, we assume a uniform label prior $p(y)$, so $p(y)$ will not affect the $\argmax$ in Eq.~(\ref{eq:predict}) and can be omitted. 

\subsection{Parameterization}
Our model, which we refer to as \gennli, is parameterized with a standard RNN-based sequence-to-sequence architecture with attention and a copy mechanism between the encoder and the decoder.\footnote{We also experimented with transformer architectures~\citep{vaswani} and found similar results.} 

\paragraph{Encoder.} Our encoder uses a standard bidirectional recurrent neural network (RNN) using long short-term memory (LSTM; \citealp{hochreiter1997long}): 

\begin{align*}
\mathbf{s}_n&= [f_{e_1}(\mathbf{v}_{n}, \overrightarrow{\mathbf{s}_{n-1}});
f_{e_2}(\mathbf{v}_{n}, \overleftarrow{\mathbf{s}_{n+1}})]
\end{align*}
\noindent 
where $f_{e_1}$ and $f_{e_2}$ are forward and backward LSTM recurrences, respectively, $\mathbf{v}_n$ is the word embedding of $x^{(p)}_n$, and $\mathbf{s}_n$ is the concatenation of the forward and backward RNN hidden states at position $n$ in the premise.

\paragraph{Decoder.} 
Our decoder uses an RNN with dot product attention from \citet{luong-etal-2015-effective} and a copy mechanism~\cite{gu-etal-2016-incorporating}. The decoder hidden state at step $t$ is computed as $\mathbf{h}_t=f_d(\mathbf{w}_t, \mathbf{h}_{t-1})$.

where $f_d$ is the forward LSTM recurrence in the decoder and $\mathbf{w}_t$ is the word embedding of $x^{(h)}_t$. 
The word distribution at position $t+1$ is computed as follows: 
\begin{align*}
{\mathbf{p}_{\vocab}}&=\softmax(\mathbf{V'}(\mathbf{V}[\mathbf{h}_t, \mathbf{s}^*_{t}, \mathbf{v}_y] + b) + b')
\end{align*}
\noindent where $\mathbf{v}_y$ is the label embedding of $y$, $\mathbf{s}_{t}^{*}$ is the context vector at step $t$ computed using attention (full details of the attention mechanism are omitted for brevity but can be found in \citealp{luong-etal-2015-effective}), and 
$\mathbf{V}$, $\mathbf{V'}$, $b$, and $b'$ are learnable parameters. Note the presence of the label embedding $\mathbf{v}_y$ 
concatenated to $\mathbf{h}_t$ and $\mathbf{s}^{*}_{t}$ to form the input to the $\softmax$ layer. This enables the label to directly influence the word distribution. We also use label-specific beginning-of-sentence (BOS) tokens as the initial symbol fed to the decoder RNN. Concretely, we create the embeddings for all BOS symbols $\bos_y$ ($y\in Y$) and prepend $\bos_{y'}$ to the hypothesis where $y'$ is the label for the instance. 

\paragraph{Copy mechanism.} 
In some datasets, hypotheses are written by humans when provided a premise and label~\citep{snli:emnlp2015}. We observed that these hypotheses sometimes appear to be written by slightly modifying the premise according to the label, e.g., adding ``not'' to negate the premise, or by replacing a phrase with a phrasal hypernym, such as replacing ``soccer game'' with ``sport'' \citep{marelli-etal-2014-sick,snli:emnlp2015}. The tokens in a premise/hypothesis pair often show a large degree of overlap. 
So we use a copy mechanism \cite{gu-etal-2016-incorporating} to (1) reduce the difficulty of word prediction when training sequence-to-sequence models on small datasets and (2) encourage the model to pay more attention to the token differences between the textual input of the encoder and decoder. We compute:  
\begin{equation}
p_{\mcopy}=\sigma (\mathbf{w}_{\mcopy}^\top [\mathbf{h}_t,\mathbf{s}^*_t,\mathbf{v}_y] + b_{\mcopy})
\end{equation}
where $p_{\mcopy} \in [0, 1]$ is the probability of copying a word from the input sequence, the vector $\mathbf{w}_{\mcopy}$ and scalar $b_{\mcopy}$ are learnable parameters, and $\sigma$ represents the logistic sigmoid function. 
We use an extended vocabulary for a specific sentence pair which includes all the words appearing in the input sentence so that the decoder can copy specific words from the input sentence instead of generating out-of-vocabulary (OOV) words.

\begin{table*}[t]
\setlength{\tabcolsep}{3pt}
\centering
\begin{tabular}{rl} 
perceptron loss:
& $\displaystyle - \log p(x^{(h)}\mid x^{(p)},y) + \max_{y'\in Y} \, \log p(x^{(h)}\mid x^{(p)},y') $   \\ 
hinge loss: &
$\displaystyle - \log p(x^{(h)}\mid x^{(p)},y) + \max_{y'\in Y} \,\{\log p(x^{(h)}\mid x^{(p)},y') + \cost(y,y')\}$     \\ 
log loss: & 
$\displaystyle - \log p(x^{(h)}\mid x^{(p)},y) + \log \sum_{y'\in Y} p(x^{(h)}\mid x^{(p)},y') $ \\
softmax-margin: & 
$\displaystyle - \log p(x^{(h)}\mid x^{(p)},y) + \log \sum_{y'\in Y} \exp \{\log p(x^{(h)}\mid x^{(p)},y') + \cost(y, y')\}$ \\
Bayes risk: & 
$\displaystyle \mathbb{E}_{p(y'\mid x^{(h)},x^{(p)})}[\cost(y, y')] = \sum_{y'\in Y}  \cost(y, y') 
\frac{p(x^{(h)}\mid x^{(p)},y')}{\sum_{y''\in Y} p(x^{(h)}\mid x^{(p)},y'')}$ \\
\theloss loss:  & 
$\displaystyle - \log p( x^{(h)} \mid x^{(p)}, y ) + \log \sum_{y' \in Y, y' \neq y} p(x^{(h)} \mid x^{(p)}, y')$
\\
\end{tabular}
\caption{Discriminative objectives considered for fine-tuning \gennli in this paper. Each is defined for a single training example $\langle x^{(p)}, x^{(h)}, y\rangle$, where $x^{(p)}$ is the premise, $x^{(h)}$ is the hypothesis, and $y\in Y$ is the label. 
}
\label{table: disc_loss_s}
\end{table*}

\section{Discriminative Fine-Tuning}

\citet{lewis2018generative} showed that generative classifiers for question answering can be improved by a discriminative fine-tuning step after estimating the generative classifier distributions. They used log loss as their discriminative objective. We also consider using a discriminative fine-tuning step when training our model, specifically we compare log loss to four other discriminative losses:
\vspace{-0.1cm}
\begin{itemizesquish}
\item \textbf{Perceptron loss}: the loss function underlying the perceptron algorithm \cite{rosenblatt1958perceptron}
\item \textbf{Hinge loss}: the loss function underlying support vector machines (SVMs) and structured SVMs \cite{wahba1999support, NIPS2003_2397}
\item \textbf{Softmax-margin}: which combines log loss with a cost function as in hinge loss  \cite{povey-08,gimpel-smith-2010-softmax}
\item \textbf{Bayes risk}: the expectation of the cost function with respect to the model's conditional distribution \cite{kaiser-00,smith-eisner-2006-minimum}
\end{itemizesquish}
Table \ref{table: disc_loss_s} shows these discriminative losses.\footnote{Again, the label prior $p(y)$ ends up canceling out because it is uniform over labels, so we do not show it.} Some losses use a \textbf{cost function}, which can be chosen by the practitioner to penalize different errors differently. In our experiments, we define it as $\cost(y,y')=1$ for $y \neq y'$ and  $\cost(y,y')=0$ if $y=y'$, where $y$ is the gold label and $y'$ is a candidate label. 

In addition, we introduce a very simple loss that is inspired by these other discriminative losses while performing quite well overall in our experiments. We call it the \textbf{\theloss loss} and define it as follows:
\begin{align}
- \! \log p( x^{(h)} \mid x^{(p)}, y ) \! + \! \log \!\sum\limits_{\substack{y' \in Y\\ y'\neq y}} \! p(x^{(h)} \mid x^{(p)}, y')
\raisetag{1\baselineskip}
\label{eq:finetune_objective}
\end{align}
\noindent 
The \theloss loss is different from log loss in that the gold label is excluded from the sum. 
Therefore, \theloss is not bounded below by zero, unlike all other discriminative losses we consider. It does not approach zero as the model becomes increasingly confident in the correct classification, as is the case with log loss and softmax-margin. Rather, \theloss is unbounded, causing learning to continually seek to increase the score of the correct label and decrease the score of the incorrect labels. 

We can view \theloss as softmax-margin with a cost function that returns $-\infty$ when $y=y'$ and 0 otherwise. However, the convention usually assumed when defining cost functions for softmax-margin is for the cost function to be nonnegative~\citep{gimpel-smith-2010-softmax}, and similar conventions are assumed with hinge loss. So we choose to use a distinct name for this loss. 

Our results in Section \ref{subsection:abaltion} show that fine-tuning using \theloss or one of the investigated discriminative losses leads to better performance than log loss fine-tuning, which was proposed for generative classifiers by \citet{lewis2018generative}. 

Though the above objectives appear discriminative due to their direct penalization of incorrect labels, they do so by using the key building blocks of generative classifiers. Thus, this fine-tuning achieves some of the benefits of discriminative classifiers while retaining the advantages of generative classifiers, as shown for question answering by \citet{lewis2018generative} and also shown in our experiments below.

\section{Experiments}
\pdfoutput=1
\label{experiment}
\subsection{Datasets}
We experiment with five sentence pair datasets, namely the Stanford Natural Language Inference corpus (SNLI; \citealp{snli:emnlp2015}), 
the SICK dataset \cite{marelli-etal-2014-sick}, 
the Multi-Genre Natural Language Inference corpus (MultiNLI; \citealp{williams-etal-2018-broad}),   
the binary Recognizing Textual Entailment (RTE; \citealp{dagan2006pascal}) dataset from the GLUE benchmark \citep{wang2018glue}, and the Microsoft Research Paraphrase Corpus (MRPC; \citealp{dolan2004unsupervised}) also from GLUE.\footnote{While MRPC is a binary paraphrase classification task rather than an NLI or entailment task, we treat it as a binary entailment task by choosing one of the sentences arbitrarily as the premise and using the other as the hypothesis.} 
The statistics of the datasets can be found in the Appendix. 
For MultiNLI, we use the matched dev set and mismatched dev set as our validation and test sets, respectively. 
Otherwise, we use the standard train, validation, and test splits from the original papers (for SNLI and SICK) or the GLUE benchmark (for RTE and MRPC).\footnote{MRPC and RTE have no public test set, so we report their performances on the development sets.} 

\subsection{Baseline Models}
We compare our \gennli model to two baseline discriminative models, and three pretrained models as described below. 

We consider InferSent \citep{conneau-etal-2017-supervised} and ESIM \citep{chen-etal-2017-enhanced} as our discriminative baselines. InferSent uses a BiLSTM network with max pooling \citep{collobert2008unified} to learn generic sentence embeddings that perform well on several NLI tasks. ESIM has a relatively complicated network structure, including a recursive architecture of local inference modeling \citep{maccartney2009natural, parikh-etal-2016-decomposable} and inference composition. 
The pretrained models we compare to are BERT~\citep{devlin-etal-2019-bert}, RoBERTa~\citep{liu2019roberta}, and XLNet~\citep{yang2019xlnet}.

We select these models as our baselines because (1) they are open-source and are frequently used as baselines for NLI tasks in related work~\citep{peters-etal-2018-deep, williams-etal-2018-broad}, and (2) their performance is strong on standard leaderboards.\footnote{GLUE leaderboard: \url{https://gluebenchmark.com/leaderboard/}; SNLI leaderboard:  \url{https://nlp.stanford.edu/projects/snli/}}

\subsection{Training Details}
Both generative and discriminative models are initialized with GloVe pretrained word embeddings \citep{pennington-etal-2014-glove}.\footnote{All of our experiments use uncased 300-dimensional GloVe vectors trained on 6 billion tokens (\url{http://nlp.stanford.edu/data/glove.6B.zip}).} 
The word embedding dimension and the LSTM hidden state dimension are set to 300. All parameters, including the word embeddings, are updated during training. The label embedding dimensionality for \gennli is set to 100.
All the experiments are conducted 5 times with different random seeds and we report the median scores. 

\paragraph{\gennli.} The training includes two steps: the model is first trained with the generative objective only (Equation \ref{eq:train_objective}) for 20 epochs, followed by the discriminative fine-tuning objective only (one of the objectives in Table \ref{table: disc_loss_s}) for 15 epochs. 
Unless otherwise specified, we use \theloss for discriminative fine-tuning. Section \ref{subsection:abaltion} compares fine-tuning objectives.\footnote{Our implementation 
is available at \url{https://github.com/tyliupku/gen-nli}}

\paragraph{Discriminative baselines.} We run the open source code of InferSent\footnote{\url{github.com/facebookresearch/InferSent}} and ESIM.\footnote{\url{github.com/coetaur0/ESIM}} Following their implementation, training stops when the performance on the dev set does not  improve across 5 consecutive epochs or the learning rate sufficiently decays (e.g,. less than $e^{-5}$).

For both \gennli and discriminative baselines, we use the Adam~\cite{kingma2014adam} optimizer with learning rates of 0.001 and 0.1, and SGD with learning rates 0.1, 0.5, 1, and 2, and select the model with the best performance on the dev set. 

\paragraph{Pretrained baselines.} We use the Hugging Face PyTorch implementation \cite{wolf1910huggingface} of pretrained transformer~\citep{vaswani} models.\footnote{\url{github.com/huggingface/transformers}} 
BERT, XLNet, and RoBERTa are configured with `bert-base-uncased', `xlnet-base-cased', and `roberta-base', respectively. 
We use the vector at the position of the [CLS] token in the last layer as the output of pretrained models, and map the output to NLI classification with a linear transformation. We fine-tune the pretrained models on our training sets for 10 epochs. We observe that the models usually converge within the first 3-5 epochs.

\section{Results}
\pdfoutput=1
\label{results}
\subsection{Data Efficiency} 
\label{section:sample_complexity}
\begin{table}[t!]
\begin{center}
\small
\setlength{\tabcolsep}{1.8mm}{
\begin{tabular}{ccccccc}

\toprule
\multicolumn{1}{c}{}
     & 5 & 20 & 100 & 500 & 1000 & all \\
\midrule
\multicolumn{1}{c}{}& \multicolumn{6}{c}{\bf{SNLI}} \\ 

\midrule
\multicolumn{1}{c}{\nligen} & \underline{\textbf{43.5}} & \underline{\textbf{45.6}} &  \underline{\textbf{50.6}} & \underline{60.6} & 64.2 & 82.2 \\ \midrule
\multicolumn{1}{c}{InferSent} & 37.5 &	39.6 &	44.1 & 56.0 & 63.9 & 84.5\\
\multicolumn{1}{c}{ESIM}  & 38.4 &	38.6 &	46.7 & 58.2 & \underline{65.4} & \underline{87.6} \\ \midrule
\multicolumn{1}{c}{BERT} & 33.4 & 37.3 & 47.4 & 	70.1 &	78.7 &	90.6 \\
\multicolumn{1}{c}{XLNet} & 34.1 &	35.6 & 45.1 & 72.3 &	77.3 &	90.9\\
\multicolumn{1}{c}{RoBERTa}  & 35.1 & 36.0 & 49.3 & \textbf{75.9} & \textbf{82.8} &	\textbf{91.7}\\

\midrule

\multicolumn{1}{c}{}& \multicolumn{6}{c}{\bf{MNLI}} \\ \midrule

\multicolumn{1}{c}{\nligen}  & \underline{\textbf{44.1}} & \underline{\textbf{47.1}}  &  \underline{\textbf{49.0}}  & \underline{60.6}  & \underline{63.4} & 67.5 \\ \midrule
\multicolumn{1}{c}{InferSent}  & 34.1  &  33.7 &  35.2  &  44.9  &  47.9 &  70.4  \\
\multicolumn{1}{c}{ESIM} &  36.9 & 35.4  &  40.5  &  49.8  &  54.2 &\underline{76.7}  \\\midrule
\multicolumn{1}{c}{BERT} & 33.0 & 34.9 & 41.6 & 63.6 &68.5 & 83.3\\
\multicolumn{1}{c}{XLNet} & 35.6& 35.6 & 39.7 & 68.2 &74.4 & 86.3\\
\multicolumn{1}{c}{RoBERTa}  &33.2& 34.9&	42.7& \textbf{68.8} &\textbf{74.6} & \textbf{87.3}\\\midrule

\multicolumn{1}{c}{}& \multicolumn{6}{c}{\bf{SICK}} \\
\midrule

\multicolumn{1}{c}{\nligen}  & \underline{\textbf{50.6}} & \underline{\textbf{64.7}} & \underline{\textbf{68.7}} & 75.2& - & 80.4 \\ \midrule
\multicolumn{1}{c}{InferSent}  &  35.5  & 46.3& 60.2 & 73.2 &-& 83.6 \\
\multicolumn{1}{c}{ESIM}  &     34.5    & 48.4& 62.9 & \underline{75.4}&-& \underline{84.6}  \\\midrule
\multicolumn{1}{c}{BERT} &  36.7    & 56.7 & 63.6 &78.6 &-&  86.0 \\
\multicolumn{1}{c}{XLNet} & 34.1 & 55.3 & 62.3 &79.0 &-&   86.8\\
\multicolumn{1}{c}{RoBERTa}  &33.5  & 56.7  & 66.3 & \textbf{83.4}& -& \textbf{88.5} \\\midrule

\multicolumn{1}{c}{}& \multicolumn{6}{c}{\bf{RTE}} \\
\midrule

\multicolumn{1}{c}{\nligen}  &  \underline{\textbf{57.0} } &  \underline{\textbf{57.7}}  &  \underline{\textbf{59.2}}  & \textbf{\underline{60.4}} & \underline{61.4}  & \underline{62.6}\\ \midrule
\multicolumn{1}{c}{InferSent}  &  49.5  &  47.3  & 52.4 & 54.2 & 55.2  & 56.3 \\
\multicolumn{1}{c}{ESIM}  &  50.1 &  50.3  &  53.5 & 55.8  & 57.3  & 58.9 \\\midrule
\multicolumn{1}{c}{BERT} &  47.3  &  48.0  &  49.1  & 59.9 & 64.3  & 66.4\\
\multicolumn{1}{c}{XLNet} &  50.9  &  53.4  &  55.9  & 60.3 & 64.6 & 68.6\\
\multicolumn{1}{c}{RoBERTa}  & 52.7   &  53.1  &  53.8  & 59.6 & \textbf{67.8}  & \textbf{74.7}\\\midrule

\multicolumn{1}{c}{}& \multicolumn{6}{c}{\bf{MRPC}} \\
\midrule

\multicolumn{1}{c}{\nligen}  & \underline{\textbf{62.8}} & \underline{64.1} & \underline{66.2} & \underline{67.8} & 69.9 & 72.9 \\ \midrule
\multicolumn{1}{c}{InferSent}  & 52.5 & 54.6 & 58.1 & 65.1 & 70.9 & 73.1 \\
\multicolumn{1}{c}{ESIM}  &  54.1  &  54.3  &  59.7  & 64.8 & \underline{71.2}  & \underline{75.1} \\\midrule
\multicolumn{1}{c}{BERT} & 53.1 & 55.0 & 57.0 &  69.6 & 74.1  & 82.3 \\
\multicolumn{1}{c}{XLNet}&  55.3  &  64.7  &  \textbf{68.5}  & 78.7 & 82.5 & 85.2\\
\multicolumn{1}{c}{RoBERTa}  & 59.8 &  \textbf{65.3}  &  67.5  & \textbf{80.3} &  \textbf{84.4} & \textbf{87.1}\\\bottomrule

\end{tabular}
}
\end{center}
\caption{Comparison of classification accuracy of  \nligen, discriminative baselines, and pretrained baselines with various amounts of training data. Here 5/20/100/500/1000 indicates the number of training instances per class. The best result for each task and data amount is shown in bold, and the best result between \gennli and the discriminative baselines is underlined.}
\label{table:SmallSet}
\end{table}

We first empirically characterize \nligen, discriminative baselines, and pretrained baselines in terms of data efficiency. We construct smaller training sets by randomly selecting 5, 20, 100, 500, and 1000 instances per class, and then train separate models across these different-sized training sets. Table~\ref{table:SmallSet} shows the results.\footnote{SICK does not have results in the 1000 column because the `contradiction' label has only 665 instances.}

When using training sets with 100 or fewer instances per class, \nligen outperforms the pretrained baselines on all datasets except for MRPC. 
We would hope that pretrained models like BERT would produce generalized text representations that would perform well after fine-tuning with a relatively small number of examples, but here we observe that a thousand or more examples is required to outperform \nligen on most datasets. 

With small training sets, \nligen also has better performance than the other discriminative baselines, though the performance gap does shrink as the training set gets larger. The accuracies become comparable when we have 1000 instances per label. 
We also see that on the full training set, the discriminative baselines outperform \nligen, which accords with our expectations and the findings of prior work \cite{ding-gimpel-2019-latent}. 

\subsection{Training Label Noise}

To measure robustness to label noise, we construct noisy datasets by randomly flipping the labels of 10\%, 30\%, or 50\% of the training instances in the binary classification tasks. The labels of other  instances are unchanged. Evaluation is done on the original validation and test sets. 

\begin{table}[t]
\begin{center}
\small
\begin{tabular}{ccccccc}
\toprule
\multicolumn{2}{c}{Accuracy}
     & 50\%  & 30\%  & 10\%  & 0\%  \\ \midrule
\multirow{3}{*}{MRPC}    
&   InferSent  & 40.6 & 61.7 & \underline{72.2} &    \underline{73.1}    \\ 
&    RoBERTa  & \underline{66.5} & \textbf{76.8} & \textbf{85.3} &  \textbf{87.1}       \\ 
&    \nligen  & \textbf{68.5} & \underline{70.0} & 71.7 &    72.9    \\ \midrule

\multirow{3}{*}{RTE}    
&   InferSent  & 50.4 & 50.9 & 54.5 &   56.3        \\ 
&   RoBERTa  & \underline{52.0} &  \textbf{63.5}  & \textbf{76.2} & \textbf{74.7} \\
&    \nligen  & \textbf{58.8} & \underline{59.9} & \underline{59.6} &  \underline{62.6} \\ 
\bottomrule
\midrule
\multicolumn{2}{c}{MCC}
     & 50\%  & 30\%  & 10\%  & 0\%  \\ \midrule

\multirow{3}{*}{MRPC}    
&   InferSent  & -0.018 & 0.189 & \underline{0.357} &    \underline{0.379}    \\ 
&   RoBERTa  & \underline{0.000} &  \textbf{0.447}  & \textbf{0.664} & \textbf{0.707} \\
&    \nligen  & \textbf{0.214} & \underline{0.245} & 0.303 &     0.352    \\ \midrule

\multirow{3}{*}{RTE}
&   InferSent  & 0.024 & 0.111 & 0.017 &   0.129        \\
&   RoBERTa  & \underline{0.030} &  \textbf{0.266}  & \textbf{0.521} & \textbf{0.501} \\
&    \nligen  & \textbf{0.173} & \underline{0.190} & \underline{0.191} &    \underline{0.230}         \\ \bottomrule

\end{tabular}
\caption{Classification accuracy and Matthews Correlation Coefficient (MCC) when using noisy training sets. The percentages are the fractions of training instances with flipped labels. 0\% is the unchanged training set. The best result for each task and each noisy setting is shown in bold, and the second-best one is underlined. }
\label{table:noisy}
\end{center}
\end{table}

Table \ref{table:noisy} shows a comparison of \nligen, InferSent, and RoBERTa on noisy datasets. In addition, we report the value of the Matthews Correlation Coefficient (MCC) \cite{matthews1975comparison}. The value of MCC ranges from -1 to 1, with higher value indicating a better classification model. 
MCC considers all values in the confusion matrix and describes it with a single number. It is viewed as a balanced measurement when the classes are of very different sizes~\cite{boughorbel2017optimal}.

We find all of the models are robust to slight noise, as the accuracy does not drop dramatically with 10\% noisy training data. However, as we increase the proportion of the label noise, the performance of InferSent decreases more rapidly than \nligen. The results are consistent between the two metrics. It is worth noting that \nligen works better than RoBERTa under the 50\%-noisy-data setting, even though RoBERTa has much stronger performance with the unchanged training set. In other words, \nligen is more robust as the performance drops only slightly with extremely noisy training data.

In general, training deep neural networks requires abundant clean data. When dealing with potentially noisy data, it may be worthwhile to build both generative and discriminative classifiers. 

\begin{table*}[t]
\centering
\small
\begin{tabular}{ccc||cccc||cccc}
\toprule
\multirow{2}{*}{Dataset} & \multirow{2}{*}{Subsampled Label} & \multirow{2}{*}{Model}  
&\multicolumn{4}{c||}{Accuracy}&\multicolumn{4}{c}{Matthews Correlation Coefficient } \\ [0.8ex]

\multicolumn{3}{c||}{}
 & 10\%   & 20\%   & 50\%   & 100\% & 10\%         & 20\%         & 50\%         & 100\%       \\ \midrule

\multirow{6}{*}{MRPC} & \multirow{3}{*}{paraphrase} 
& InferSent    & 49.2  & 63.1  & 70.6 &    \underline{73.1}   & 0.244       & 0.362       & \underline{0.372}       &    \underline{0.379}        \\
&& RoBERTa   &   \textbf{74.1} &\textbf{83.2} & \textbf{86.0} &   \textbf{87.1}   & \textbf{0.526} & \textbf{0.645} & \textbf{0.688} &    \textbf{0.707}   \\
&& \nligen     & \underline{70.2}  & \underline{70.7}  & \underline{72.0}   &   72.9   & \underline{0.301}       & \underline{0.367}       & 0.318       &      0.352     \\ \cmidrule{2-11} 

& \multirow{3}{*}{non-paraphrase} 
& InferSent    & 68.3 & \underline{70.9} & \underline{73.8}  &   \underline{73.1}   & 0.191       & 0.287       & \underline{0.373}       &      \underline{0.379}       \\
&& RoBERTa   &  \textbf{77.2} & \textbf{81.2} & \textbf{86.3} &   \textbf{87.1}   & \textbf{0.469} & \textbf{0.568} & \textbf{0.697} &    \textbf{0.707}   \\
&& \nligen     & \underline{70.8}  & 70.3  & 72.2  &    72.9    & \underline{0.333}       & \underline{0.292}       & 0.319       &   0.352        \\ 
\midrule

\multirow{6}{*}{RTE} & \multirow{3}{*}{entailment} 
& InferSent    & 47.3  & 47.3  & 52.3  &  56.3  & 0.000		           & 0.036        & 0.135        &     0.129       \\
&& RoBERTa   & \textbf{66.7} & \textbf{66.7} & \textbf{71.5} &   \textbf{74.7}   & \textbf{0.226} & \textbf{0.230} & \textbf{0.426} &   \textbf{0.501}    \\
&& \nligen     & \underline{55.8}  & \underline{56.5}  & \underline{59.9}  &  \underline{62.6}    & \underline{0.128}        & \underline{0.135}        & \underline{0.194}        &     \underline{0.230}       \\ 
\cmidrule{2-11}

& \multirow{3}{*}{non-entailment} 
& InferSent    & 52.7  & 52.7  & 54.0    &   56.3   & 0.001           & 0.035        & 0.065        &      0.129      \\
&& RoBERTa  &   \underline{56.0}    & \textbf{62.1}  &   \textbf{72.9}    &   \textbf{74.7}    &  \underline{0.177}   &   \textbf{0.371}   &  \textbf{0.471}  & \textbf{0.501}  \\
&& \nligen     & \textbf{60.5}  & \underline{60.3}  & \underline{62.2}  &  \underline{62.6}      & \textbf{0.209}        & \underline{0.204 }       & \underline{0.181}        &      \underline{0.230}      \\ 
\midrule

& \multirow{3}{*}{entailment}
& InferSent    & 57.4 & 60.1 & \underline{67.8} &   \underline{70.4}    & 0.396       & 0.431       & \underline{0.522}       &      \underline{0.557}     \\
&& RoBERTa & \textbf{82.4} & \textbf{84.8} & \textbf{87.0} & \textbf{87.3} & \textbf{0.747} & \textbf{0.776} & \textbf{0.806} & \textbf{0.809} \\
&& \nligen     & \underline{60.8} & \underline{61.7} & 67.1 &   67.5   & \underline{0.410}       & \underline{0.452}       & 0.497       &     0.512       \\ 
\cmidrule{2-11}

\multirow{3}{*}{MNLI} & \multirow{3}{*}{neutral}
& InferSent    & 60.5 & 62.5 & \underline{68.8} &   \underline{70.4}   & 0.445       & 0.469       & \underline{0.539}       &      \underline{0.557}      \\ 
&& RoBERTa & \textbf{83.0}   & \textbf{84.5} & \textbf{85.9} & \textbf{87.3} & \textbf{0.754} & \textbf{0.769} & \textbf{0.790}  & \textbf{0.809} \\
&& \nligen     & \underline{61.7} & \underline{63.8} & 67.6 &  67.5    & \underline{0.463}       & \underline{0.487}       & 0.491       &         0.512   \\ 
\cmidrule{2-11}

& \multirow{3}{*}{contradiction}
& InferSent    & 60.8 & \underline{64.0} & \underline{67.9}  &   \underline{70.4}    & \underline{0.444}       & \underline{0.479}       & \underline{0.526}       &    \underline{0.557}        \\
&& RoBERTa & \textbf{82.7} & \textbf{84.5} & \textbf{86.6} & \textbf{87.3} & \textbf{0.748} & \textbf{0.773} & \textbf{0.800} & \textbf{0.809} \\
&& \nligen     & \underline{61.0} & 62.0 & 65.6 &  67.5    & \underline{0.444}       & 0.466       & 0.492       &       0.512     \\ 
\bottomrule
\end{tabular}
\caption{Classification accuracies and Matthews Correlation Coefficients of test sets when training on label-imbalanced training sets. 
Column headers indicate the percentage of the subsampled label's training instances that are retained in the training set. All training instances are used for the other labels. 
The best result for each task and each subsample setting is shown in bold, and the second-best one is underlined.
}
\label{table:imbalanced}
\end{table*}

\subsection{Imbalanced Label Distributions} \label{sec:Imbalanced_Label_Distributions}

We also perform experiments in a setting with label imbalance in the training set. Each imbalanced training set is constructed by random sampling and keeping only 10\%, 20\%, or 50\% of the instances from one selected class, and keeping all the instances from the other classes. We use the original validation and test sets. We still use a uniform prior for \gennli. 

Table~\ref{table:imbalanced} shows the comparison of generative, discriminative, and BERT-based classifiers under various imbalanced training sets.\footnote{We report the results on these three datasets since they represent different characteristics in terms of training set size, number of candidate labels, and performance difference between \nligen and InferSent on the full training set.} Aside from the 10\%-non-entailment RTE dataset, RoBERTa always performs the best. This is unsurprising because, even after subsampling, the training set sizes are on a similar order of magnitude as the full sets, with which RoBERTa excels (Table \ref{table:SmallSet}). However, RoBERTa does show degradation as the subsampling rate becomes more extreme (more than 10\% in MRPC, 8-18\% in RTE, and 4-5\% on MNLI). \gennli shows a smaller or comparable decrease in performance, though its overall accuracies are lower. In comparing the generative and discriminative classifiers, \nligen always outperforms InferSent when keeping only 10\% of the instances for the selected class. However, as the percentage of instances in the selected class increases, InferSent begins to perform better than \nligen.

Another finding is that the different labels have different effects under the imbalanced setting. For example, the performance of RTE/non-entailment decreases more slowly than RTE/entailment for both \nligen and InferSent, which might suggest that the non-entailment label requires fewer training examples than entailment. 

Data efficiency might also affect performance under the label imbalanced setting. We believe it is not the only factor for a performance difference between the generative and discriminative models, as the MNLI dataset has 130k instances per class and the training set still has more than 270k instances in total even under the 10\% setting, indicating \nligen has certain advantages over InferSent when the label distribution is imbalanced.

\section{Analysis}
\pdfoutput=1
\label{analysis}
\label{subsection:abaltion}

\begin{table}[t]
\begin{center}
\small
\begin{tabular}{p{5.1cm}cc}
\toprule
\multicolumn{1}{c}{}
     & SNLI & RTE  \\ \midrule
\nligen  &  82.2 &   62.6 \\ 
\quad no copy mechanism  & 74.4 &  54.7   \\ 
\quad no generative training  &  80.1   & 60.3        \\
\quad no discriminative fine-tuning  &  79.1  &  61.7  \\
\midrule 
\nligen, $p(x^{(h)} \mid x^{(p)}, y)$ & 82.2 & 62.6 \\
\nligen, $p(x^{(p)} \mid x^{(h)}, y)$ & 77.1 & 59.7 \\
\bottomrule
\end{tabular}
\end{center}
\caption{Results showing contribution of individual  modeling/training decisions on SNLI and RTE.}
\label{table:ablationstudy}
\end{table}

\subsection{Modeling and Training Decisions} 
We now empirically assess the importance of major components of modeling and training. As shown in Table \ref{table:ablationstudy}, the copy mechanism is essential, which meets our expectation because we observe a lot of lexical overlap between the premise and hypothesis in many pairs.\footnote{All the experiments in our paper are in-domain testing. We also test \gennli in out-of-domain (OOD) datasets to see whether the copy mechanism is helpful in this case. For example, we train on MNLI and test on SICK. The trend is not consistent across different OOD settings.} We find both generative training and fine-tuning objectives to be helpful, as better results are achieved by training with both objectives. 

\gennli defines the conditional distribution of hypotheses given a premise and label. We could instead model $p(x^{(p)} \mid x^{(h)}, y)$. 
The final two rows of Table \ref{table:ablationstudy} compare the two, showing  better performance with $p(x^{(h)} \mid x^{(p)}, y)$. The difference is larger in SNLI, which may be due in part to how the dataset was created. If annotators are provided with a premise and label and asked to write hypotheses, as in SNLI, we would expect that a generative model that matches this process would excel. The difference may also be due to the fact that in the entailment pairs, the premise often has more information than the hypothesis, and it is expected to be easier to remove information (when generating the hypothesis from the premise) than to add it.

\subsection{Discriminative Fine-Tuning Comparison} 

Table \ref{table:disc_obj_results} compares discriminative fine-tuning objectives.\footnote{Note that all models are trained with the generative objective before discriminative fine-tuning. Results for other datasets are provided in the Appendix.} 
Several choices, including hinge, softmax-margin, and \theloss, consistently outperform the log loss used as discriminative fine-tuning objective by \citet{lewis2018generative}. The perceptron loss and Bayes risk also often outperform log loss. It is worth noting that \theloss performs the best when using the full training set on four out of five datasets (see Appendix for full results), while softmax-margin is best with smaller training sets. 
These results suggest that improving discriminative fine-tuning does not harm the data efficiency benefits of generative classifiers, but rather is able to accentuate them.

\begin{table}[t]
\small
\addtolength{\tabcolsep}{-1.5pt}
\begin{tabular}{lccc|ccc}
\toprule
\multicolumn{1}{c}{}&\multicolumn{3}{c|}{SNLI}&\multicolumn{3}{c}{RTE} \\
           & 100  & 1000  &  all  & 100  & 1000  &  all \\
\midrule
perceptron               & 49.6             & 62.5             & 80.4              & 57.9             & 60.1             & 61.1             \\
hinge                    & 49.9             & 63.1             & 81.1             & \underline{58.8} & \underline{61.3}             & \underline{62.2} \\
log                      & 49.1             & 62.3             & 80.7             & 57.4             & 59.7             & 60.5             \\
softmax-margin               & \textbf{50.6}    & \textbf{64.2}    & \underline{81.9}    & \textbf{59.2}    & 61.1             & \underline{62.2} \\
\theloss                 & \underline{50.0}   & \underline{63.7} & \textbf{82.2}    & 58.1             & \textbf{61.4}    & \textbf{62.6}    \\
Bayes risk                     & 49.0               & 62.6             & 80.1             & 58.3             & 60.6             & 61.4             \\
\bottomrule
\end{tabular}
\caption{Comparision of discriminative fine-tuning objectives on SNLI and RTE datasets. The best result for each task and data amount is shown in bold, and the second-best one is underlined. }
\label{table:disc_obj_results}
\end{table}

\subsection{Data Generation}

\begin{table}[t]
\centering
\small
\setlength{\tabcolsep}{2pt}
\begin{tabular}{ccp{6.5cm}}
\toprule
\multicolumn{3}{c}{\textbf{\gennli trained on full SICK training set}} \\
\midrule

\multirow{5}{*}{N} & 
$x^{(p)}$ &   A man is sitting near a bike and is writing a note. \\
& $x^{(h)}$ &   A man with paint covered clothes is sitting outside in a busy area writing something. \\
& gen. &   \textbf{A man is sitting in a bike and is writing a note in a busy area.} \\
\midrule

\multirow{5}{*}{E} &  
$x^{(p)}$ &   People wearing costumes are gathering in a forest and are looking in the same direction. \\
& $x^{(h)}$ &   Masked people are looking in the same direction in a forest. \\
& gen. &   \textbf{People wearing costumes are looking in a forest.} \\
\midrule

\multirow{5}{*}{C} & 
$x^{(p)}$ &   There is no child holding a water gun or getting sprayed with water. \\
& $x^{(h)}$ &   A laughing child is holding a water gun and getting sprayed with water. \\
& gen. &   \textbf{A child is holding a water gun.} \\

\bottomrule
\toprule
\multicolumn{3}{c}{\textbf{\gennli trained on small SICK training set}} \\
\midrule

\multirow{5}{*}{N} & 
$x^{(p)}$ &  A little girl and a woman wearing a yellow shirt are getting splashed by a city fountain.  \\
& $x^{(h)}$ &  The young girl is playing on the edge of a fountain and an older woman is watching her.  \\
& gen. &   \textbf{A little girl is playing in the background.} \\

\midrule

\multirow{3}{*}{E} &  
$x^{(p)}$ &   A man is playing a flute. \\
& $x^{(h)}$ &   A man is playing the flute. \\
& gen. &   \textbf{A flute is being played by a man.} \\

\midrule

\multirow{5}{*}{C} & 
$x^{(p)}$ &   There is no man on a rock high above some trees standing in a strange position. \\
& $x^{(h)}$ &   A man is on a rock high above some trees and is standing in a strange position. \\
& gen. &   \textbf{A man is on a rock high above some trees is standing in a strange position.} \\

\bottomrule
\end{tabular}
\caption{Generated hypotheses for premises with given labels (N = neutral, E = entailment, C = contradiction).} 
\label{table:generation-small}
\end{table}

One advantage of generative models is that they can be used to generate samples in order to interpret how the model works. Since we include label information in the decoder of \nligen, we are able to generate various hypotheses for a premise by  specifying the label. 
Table \ref{table:generation-small} shows example generations from two models, one using the full dataset for training and the other using a small training set with only 500 examples per class. We use greedy decoding for these generations. 

We observe that the generated examples comport with the labels and premises we have specified, and the generation is of high quality in terms of fluency. 
However, the diversity is relatively low, with the generated samples looking similar to the premise. This is not surprising since we assume the decoder relies heavily on the copy mechanism when trained on NLI pairs, as some hypotheses differ only slightly  from their corresponding premises. The generations are relatively short compared to the gold hypotheses, which is likely due in part to greedy decoding. 
The model might require more training data and/or a different decoding algorithm to be able to produce more diverse generations. 
We also note that generations for the entailment label generally look better than those for contradiction.\footnote{Future work may consider using these generations for data augmentation. While our preliminary experiments in this direction were not positive, future work will consider fine-tuning pretrained language models as generative classifiers and using them with diverse decoding strategies to automatically expand small training sets.}

\section{Conclusions and Future Work}
\pdfoutput=1
\label{conclusion}

We proposed \gennli, a discriminatively-finetuned generative classifier for NLI tasks, and empirically characterized its performance by comparing it to  discriminative models and pretrained models. We found several discriminative fine-tuning objectives to outperform log loss, including \theloss, a simple but effective choice. We conducted extensive experiments with  \gennli, showing its robustness across challenging empirical conditions. We also showed its ability to generate hypotheses given premises and particular labels. Future work may explore generating of diverse sets of hypotheses for a given premise and label, with the goal of performing data augmentation. Other future work will be to measure the performance of \gennli on adversarial and similarly challenging NLI datasets.

\section*{Acknowledgments}
We would like to thank Sam Wiseman for contributions to an earlier version of this manuscript, and the anonymous reviewers for their helpful feedback. Z.~Sui and B.~Chang thank NSFC (No. 61876004, No. U19A2065) and Beijing Academy of Artificial Intelligence (BAAI) for their generous support.

\bibliography{emnlp2020,anthology}
\bibliographystyle{acl_natbib}

\appendix
\section{Appendix}
\pdfoutput=1
\subsection{Dataset}

We present our results on the five publicly available NLI datasets shown in Table \ref{table:data_statistics}, which 
include the Stanford Natural Language Inference (SNLI) corpus \cite{snli:emnlp2015},  the SICK corpus \cite{marelli-etal-2014-sick}, the Multi-Genre Natural Language Inference corpus (MultiNLI) \cite{williams-etal-2018-broad}, the Recognizing Textual Entailment (RTE) \cite{dagan2006pascal} corpus,  and the Microsoft Research Paraphrase Corpus (MRPC) from the GLUE benchmark \cite{wang2018glue}.\footnote{For the corpora with no public test set, we report the performance on the dev set in our paper.} 
For MultiNLI, we use the matched dev set and mismatched dev set as our validation and test sets, respectively. Table \ref{table:data_statistics} shows the statistics of the datasets in our paper. We use the standard train, validation, and test divisions from the original papers (SNLI, MultiNLI and SICK) or GLUE benchmark (RTE and MRPC).
These datasets can be downloaded at \url{https://nlp.stanford.edu/projects/snli/},  \url{https://gluebenchmark.com}, and \url{http://marcobaroni.org/composes/sick.html}.

\begin{table}[t]
\centering
\small
\begin{tabular}{ccccc}
\toprule
Dataset & \#Train & \#Valid & \#Test & \#Class  \\\midrule
SNLI & 549K & 9.8K & 9.8K & 3\\ 
MultiNLI & 392K & 9.8K & 9.8K & 3\\
SICK & 4.5K & 0.5K & 4.9K & 3\\
RTE & 2.4K & 0.2K & - & 2\\
MRPC & 4.0K & 1.7K & - & 2\\\bottomrule
\end{tabular}
\caption{Dataset statistics.}
\label{table:data_statistics}
\end{table}

\begin{table}[t]
\setlength{\tabcolsep}{4pt}
\small
\centering 
\begin{tabular}{lllllll}
\toprule
           & 5    & 20   & 100  & 500  & 1000 & all  \\
\midrule
\multicolumn{7}{c}{\textbf{SNLI}}                             \\ \midrule
perceptron & 41.8 & 44.1 & 49.6 & 58.4 & 62.5 & 80.4 \\
hinge      & 42.3 & \underline{45.3} & 49.9 & 58.6 & 63.1 & 81.1 \\
log        & 42.1 & 43.2 & 49.1 & 58.6 & 62.3 & 80.7 \\
softmax-margin & \textbf{43.5} & \underline{45.3} & \textbf{50.6} & \textbf{60.6} & \textbf{64.2} & \underline{81.9} \\
\theloss & 42.7 & \textbf{45.6} & \underline{50.0}   & \underline{59.8} & \underline{63.7} & \textbf{82.2} \\
Bayes risk    & \underline{42.8} & 44.7 & 49.0   & 58.3 & 62.6 & 80.1 \\
\midrule
\multicolumn{7}{c}{\textbf{MNLI}}       \\ \midrule
perceptron & 42.7 & 45.5 & 46.7 & 58.1 & 61.6 & 66.3 \\
hinge      & \underline{43.2} & \underline{46.3} & \underline{48.2} & \underline{60.2} & \underline{62.8} & 67.1 \\
log        & 42.1 & 45.4 & 46.7 & 58.3 & 61.4 & 66.2 \\
softmax-margin & \textbf{44.1} & \textbf{47.1} & \textbf{49.0}   & \textbf{60.6} & \textbf{63.4} & \textbf{67.5} \\
\theloss & 42.3 & 45.9 & 47.7 & 60.0   & \underline{62.8} & \underline{67.3} \\
Bayes risk    & 43.1 & 45.7 & 47.7 & 59.1 & 61.6 & 66.2 \\
\midrule
\multicolumn{7}{c}{\textbf{SICK}}     \\ \midrule
perceptron & 49.1 & 61.7 & 66.9 & 73.4 &   -   & 79.7 \\
hinge      & \textbf{50.6} & \underline{63.8} & 67.8 & 73.6 &   -   & 80.0 \\
log        & 48.6 & 62.1 & 67.5 & 73.1 &  -    & 79.8 \\
softmax-margin & \underline{50.2} & \textbf{64.7} & \textbf{68.7} & \underline{74.3} &  -    & \underline{80.2} \\
\theloss & 48.4 & 62.4 & \underline{68.3} & \textbf{75.2} &   -   & \textbf{80.4} \\
Bayes risk     & 48.2 & 62.4 & 67.2 & 72.8 & - & 79.7 \\
\midrule
\multicolumn{7}{c}{\textbf{RTE}}     \\ \midrule
perceptron & 56.1 & \underline{57.4} & 57.9 & 59.4 & 60.1 & 61.1 \\
hinge      & 56.4 & 57.1 & \underline{58.8} & 59.2 & \underline{61.3} & \underline{62.2} \\
log        & 56.5 & 57.1 & 57.4 & 59.1 & 59.7 & 60.5 \\
softmax-margin & \textbf{57.0}   & \textbf{57.7} & \textbf{59.2} & \textbf{60.4} & 61.1 & \underline{62.2} \\
\theloss & \underline{56.7} & \underline{57.4} & 58.1 & \underline{59.6} & \textbf{61.4} & \textbf{62.6} \\
Bayes risk     & 56.1 & 57.2 & 58.3 & 59.3 & 60.6 & 61.4 \\
\midrule
\multicolumn{7}{c}{\textbf{MRPC}}     \\ \midrule
perceptron & 62.1 & 62.5 & 64.6 & 66.1 & 68.6 & 69.8 \\
hinge      & 62.3 & \underline{63.8} & 65.4 & 67.1 & 69.0   & 71.8 \\
log        & 61.7 & 62.1 & 64.1 & 65.9 & 68.1 & 71.3 \\
softmax-margin & 62.6 & \textbf{64.1} & \textbf{66.2} & \textbf{67.8} & \textbf{69.9} & \underline{72.8} \\
\theloss & \underline{62.8} & 63.7 & \underline{65.6} & 67.4 & \underline{69.8} & \textbf{72.9} \\
Bayes risk     & \textbf{63.2} & 63.5 & \underline{65.6} & \underline{67.7} & 69.5 & 72.5 \\
\bottomrule
\end{tabular}
\caption{Comparison of discriminative fine-tuning objectives. The best result for each task and data amount is shown in bold, and the second-best one is underlined. 
}
\label{table:disc_obj}
\end{table}

\subsection{Discriminative Fine-Tuning Comparison} 
Table \ref{table:disc_obj} lists the full comparison results of different discriminative fine-tuning objectives.
Several choices, including hinge, softmax-margin, and \theloss, consistently outperform the log loss used as discriminative fine-tuning objective by \citet{lewis2018generative}. It is worth noting that \theloss performs the best when using the full training set on four out of five datasets.

\subsection{Data Generation}
Table \ref{table:generation-RTE} shows example generations from two models, one using the full dataset for training and the other using a small training set with only 500 examples per class. 

\begin{table}[t]
\centering
\small
\setlength{\tabcolsep}{2pt}
\begin{tabular}{ccp{6.5cm}}
\toprule
\multicolumn{3}{c}{\textbf{\gennli trained on full RTE training set}} \\
\midrule

\multirow{5}{*}{E} & 
$x^{(p)}$ &    Only a week after it had no comment on upping the storage capacity of its hotmail e-mail service , microsoft early thursday announced it was boosting the allowance to 250mb to follow similar moves by rivals such as google , yahoo , and lycos. \\
& $x^{(h)}$ &   Microsoft 's hotmail has raised its storage capacity to 250mb. \\
& gen. &   \textbf{Microsoft was boosting of its hotmail e-mail.} \\
\midrule

\multirow{5}{*}{N} &  
$x^{(p)}$ &   The name for the newest james bond film has been announced today . the 22nd film , previously known only as `` bond 22 '' , will be called `` quantum of solace '' . Eon productions who are producing the film made the announcement today at pinewood studios , where production for the film has been under way since last year . The name of the film was inspired by a short story of the same name from for your eyes only by bond creator , ian fleming. \\
& $x^{(h)}$ &   James bond was created by ian fleming. \\
& gen. &   \textbf{James bond is a member of the film.} \\

\bottomrule
\toprule
\multicolumn{3}{c}{\textbf{\gennli trained on small RTE training set}} \\
\midrule

\multirow{5}{*}{E} & 
$x^{(p)}$ &  Lin piao , after all , was the creator of mao 's `` little red book '' of quotations.  \\
& $x^{(h)}$ &  Lin piao wrote the `` little red book '' .  \\
& gen. &   \textbf{Lin piao 's `` little red book '’.} \\

\midrule

\multirow{3}{*}{N} &  

$x^{(p)}$ &  A dog is pushing a toddler into a rain puddle. \\
& $x^{(h)}$ &  A dog is pulling a toddler out of a rain puddle. \\
& gen. &   \textbf{A dog is pushing a rain puddle.} \\
\bottomrule
\end{tabular}
\caption{Generated hypotheses for premises with given labels (N = not entailment, E = entailment).} 
\label{table:generation-RTE}
\end{table}

\begin{table*}[t]
\centering
\small
\setlength{\tabcolsep}{3pt}
\begin{tabular}{ccp{11.5cm}}
\toprule
\multirow{5}{*}{Neutral} & 
$x^{(p)}$ &    A brown dog is attacking another animal in front of the man in pants. \\
& $x^{(h)}$ &   Two dogs are fighting. \\
& gen. &   \textbf{A brown dog is attacking a brown dog in front of the man.} \\
& gen. w/ finetune &   \textbf{A man is sitting on a black shirt is standing on a black shirt.} \\
& gen. w/o copy &   \textbf{A man is wearing a black shirt and is sitting on a dirt ball.} \\
\midrule

\multirow{6}{*}{Entailment} &  
$x^{(p)}$ & A group of children in uniforms is standing at a gate and one is kissing the mother. \\
& $x^{(h)}$ & A group of children wearing the same clothes is waiting at a gate and one is kissing the mother \\
& gen. &   \textbf{A group of children in uniforms is standing at a gate.} \\
& gen. w/ finetune &   \textbf{A group in uniforms at uniforms is gate and one is kissing mother.} \\
& gen. w/o copy &   \textbf{A man is sitting on a ball in the water.} \\
\midrule

\multirow{5}{*}{Contradiction} &  
$x^{(p)}$ & There is no child holding a water gun or getting sprayed with water. \\
& $x^{(h)}$ & A laughing child is holding a water gun and getting sprayed with water. \\
& gen. &   \textbf{A child is holding a water gun.} \\
& gen. w/ finetune &   \textbf{There is child child holding a water gun with water.} \\
& gen. w/o copy &   \textbf{A dog is jumping in the water.} \\
\bottomrule

\end{tabular}
\caption{Generated hypotheses for premises with given labels using models trained on the full SICK dataset. When generating using the discriminatively-finetuned model, the outputs show more repetition, while without the copy mechanism, they drift more from the premise.} 
\label{table:copy-ablation}
\end{table*}

\subsection{Ablation of Copy Mechanism in Generation}
Table \ref{table:copy-ablation} shows the generated hypotheses of the proposed generative classifier. Comparing the generative classifiers with and without copy mechanism, we find that the copy mechanism can help the model capture key differences between premise and hypothesis sentences given the specified labels. For example, we see `There is no child' versus `A child' given the label `contradiction', and `another animal' versus `a brown dog' given the label `neutral'. The copy mechanism also helps to avoid excessive semantic drift, e.g., generating the same subject as the premise and maintaining a reasonable amount of text with the premise. 

Although classification accuracy increases by adopting discriminative finetuning after generative training, the finetuning method can lead to ungrammatical or repetitive generated sentences, as demonstrated in Table \ref{table:copy-ablation}. This shows that generated text with higher quality does not necessarily lead to better performance in NLI classification.

\end{document}